%% file: 2020_MICCAI_ActiveContour.tex
\documentclass[runningheads,a4paper]{llncs}

\usepackage{amssymb}
\usepackage[utf8]{inputenc}
\usepackage{graphicx}
\usepackage{amsmath}
\usepackage{xcolor}
\usepackage{subfigure}
\usepackage{wrapfig}
\usepackage[linesnumbered,ruled,vlined]{algorithm2e}

\SetKwInput{KwInput}{Input}                
\SetKwInput{KwOutput}{Output}

\usepackage{url}
\urldef{\mailsa}\path|{alfred.hofmann, ursula.barth, ingrid.haas, frank.holzwarth,|
\urldef{\mailsb}\path|anna.kramer, leonie.kunz, christine.reiss, nicole.sator,|
\urldef{\mailsc}\path|erika.siebert-cole, peter.strasser, lncs}@springer.com|

\newcommand{\RN}[1]{%
	\textup{\lowercase\expandafter{\it \romannumeral#1}}%
}

\newcommand{\etal}{\textit{et al}.}

\begin{document}

\mainmatter  

\title{An End-to-End learnable Flow Regularized Model for Brain Tumor Segmentation}

\titlerunning{Flow Regularized Model for Brain Tumor Segmentation}

%
%
\author{Yan Shen,\ Zhanghexuan Ji\ and Mingchen Gao }
\authorrunning{Y. Shen \etal}

\institute{
Department of Computer Science and Engineering,
University at Buffalo,\\
The State University of New York, Buffalo, USA\\
}

%
%

\toctitle{Flow Regularized Model for Brain Tumor Segmentation}
\maketitle

\begin{abstract}
Many segmentation tasks for biomedical images can be modeled as the minimization of an energy function and solved by a class of max-flow and min-cut optimization algorithms. However, the segmentation accuracy is sensitive to the contrasting of semantic features of different segmenting objects, as the traditional energy function usually uses hand-crafted features in their energy functions. To address these limitations, we propose to incorporate end-to-end trainable neural network features into the energy functions. Our deep neural network features are extracted from the down-sampling and up-sampling layers with skip-connections of a U-net. In the inference stage, the learned features are fed into the energy functions. And the segmentations are solved in a primal-dual form by ADMM solvers. In the training stage, we train our neural networks by optimizing the energy function in the primal form with regularizations on the min-cut and flow-conservation functions, which are derived from the optimal conditions in the dual form. We evaluate our methods, both qualitatively and quantitatively, in a brain tumor segmentation task. As the energy minimization model achieves a balance on sensitivity and smooth boundaries, we would show how our segmentation contours evolve actively through iterations as ensemble references for doctor diagnosis. 

\end{abstract}

\input{./intro}

\input{./method_v1}

\input{./exp}

\vspace{-1em}
\section{Conclusion}
We propose an active contour model with deep U-Net extracted features. Our model is trainable end-to-end on an energy minimization function with flow-regularized optimal constraints. In the experiments, we show that the performance of our methods is comparable with state-of-the-art. And we also demonstrate how the segmentation evolves with the number of iterations and level set thresholds.  


%
%

\bibliographystyle{abbrv}
\bibliography{2020_MICCAI_ActiveContour}

\end{document}

%% file: intro.tex
\section{Introduction}
Brain Tumors are fatal diseases affecting more than 25,000 new patients every year in the US. Most brain tumors are not diagnosed until symptoms appear, which would significantly reduce the expected life span of patients. Having early diagnosis and access to proper treatments is a vital factor to increase the survival rate of patients. 
MRI image has been very useful in differentiating sub-regions of the brain tumor. 
Computer-aided automatic segmentation distinguishing those sub-regions  would be a substantial tool to help diagnosis. 

Among these automatic medical imaging algorithms, the advent of deep learning is a milestone. These multi-layer neural networks have been widely deployed on brain tumor segmentation systems. Chen \etal ~\cite{chen2018mri} uses a densely connected 3D-CNN to segment brain tumor hierarchically. Karimaghaloo \etal ~\cite{karimaghaloo2016adaptive} proposes an adaptive CRF following the network output to further produce a smooth segmentation.  Qin \etal ~\cite{qin2018autofocus} and Dey \etal ~\cite{dey2018compnet} use attention model as modulations on network model. Zhou \etal ~\cite{zhou2018one} uses transfer learning from different tasks. Spizer \etal ~\cite{spitzer2018improving} and Ganaye \etal ~\cite{ganaye2018semi} use spatial correspondence among images for segmentation. Le \etal ~\cite{le2018deep} proposes a level-set layer as a recurrent neural network to iteratively minimize segmentation energy functions. These methods combine the benefit of low inductive bias of deep neural network with smooth segmentation boundaries. 

One fundamental difficulty with distinguishing tumor segmentation is that their boundaries have large variations. For some extreme conditions, even experienced doctors have to vote for agreements for a final decision. Under those circumstances, a unique deterministic output of segmentation result is insufficient. Active contour models (ACMs), which is firstly proposed by Kass \etal ~\cite{kass1988snakes}, are widely applied in biomedical image segmentation before the era of deep learning. ACMs treat segmentation as an energy minimization problem. They are able to handle various topology changes naturally by providing multiple optimal solutions as a level set with different thresholds. In the past two decades, quite a number of variations of ACMs have been proposed, such as active contour without edge~\cite{chan2001active}, with balloon term~\cite{cohen1991active}. Yuan \etal~\cite{yuan2010study} proposes a flow based track for deriving active segmentation boundaries. The evolving boundaries provide a coarse to fine separation boundary from the background. 

In this paper, we present a trainable deep network based active contour method. Deep neural network has tremendous advantages in providing global statistics at semantic level. However, deep features do not preserve the boundary geometry, such as topology and shapes. In our model, we explicitly use the active contour to model the boundary and optimize its flow information. 
Different from traditional active contour methods, the energy function takes neural network trained features. The energy minimization problems are reformulated as a max-flow/min-cut problem by introducing an auxiliary dual variable indicating the flow transportation. We link the optimal segmentation conditions with the saturated flow edges with minimal capacities. By utilizing this property, we design a trainable objective function with extra flow regularization terms on these saturated edges. Our work combines the benefit of deep neural network and active contour models. 

Our contributions can be summarized as follows: (1) We present a trainable flow regularized loss combined with neural network architecture, leveraging the benefits of feature learning from neural networks and the advantages of regularizing geometry from active contour models. (2) Our results achieve comparable results with state-of-the-art, and more importantly, show that they are flexible to be adjusted to meet the ambiguity nature of biomedical image segmentation. 

%% file: method_v1.tex
\vspace{-0.1in}
\section{Methods}
\subsection{Image Segmentation as Energy Minimization Problem}
Chan \etal ~\cite{chan2006algorithms} considered image segmentation with minimizing the following energy functions
\begin{equation}
\label{eq1}
\min_{\lambda(x)\in \{0,1\}} \int_{\Omega} (1 - \lambda(x))C_{s}(x)dx +\int_{\Omega}   \lambda(x)C_{t}(x)dx + \int_{\Omega} C_g(x) |\nabla\lambda(x)|dx
\end{equation}
where $C_{s}(x)$ is foreground pixels, $C_t(x)$ is background pixels and $C_g(x)$ is edge pixels. As a naive treatment, $C_{s}(x)$ and $C_t(x)$ could use a simple thresholding of original image, $C_g(x)$ could use a filtered version of original image by Sobel edge detectors.\\ 
\textbf{Relation to Level Set}
Chan \etal ~\cite{chan2006algorithms} constructs a relationship between the global optimum of the binary relaxed problem of $\lambda(x) \in[0 ,1]$ with the original binary problem (\ref{eq1}). Specifically, let $\lambda^*(x)$ be a global optimum of (\ref{eq1}), its thresholding $\lambda^{l}(x)$ defined as
\begin{equation}
\lambda^l(x) =
\begin{cases}
0     & \quad \text{when} \lambda^*(x) > l \\
1  & \quad \text{when} \lambda^*(x) \leq l
\end{cases}
\end{equation}
is a global solution for (\ref{eq1}) with any $l\in [0,1]$. The function $\lambda^{l}(x)$ indicates the level set $S^{l}$.
\subsection{Reformulation as Max-Flow/Min-Cut Problem}
The above energy minimization model for image segmentation could be transformed as the well-known max-flow/min-cut~\cite{greig1989exact} problems   
\begin{align}
\max_{p_s, p_t, \mathbf{p}}\min_{\lambda(x)\in \{0,1\}} & \int_{\Omega} p_{s}(x)dx +
\int_{\Omega}  \lambda(x)(divp(x) - p_s(x) + p_t(x))dx
\nonumber\\
\text{s.t} \hspace{1em}  & p_s(x) \leq C_s(x), p_t(x) \leq C_t(x) , |p| \leq C_g(x) \label{eq10} 
\end{align}
where $p_s$, $p_t$ and $divp$ are inflow, outflow and edgeflow capacity with constraints up to the term $C_s$, $C_t$ and $C_g$ in energy minimization functions. The correlations with flow model are shown in Fig. \ref{fig:flow}. We give a  ADMM~\cite{mangasarian1994nonlinear} solution in Algorithm \ref{al1}.\\
\begin{algorithm}[H]
	\label{al1}
	\KwInput{$C_s(x), C_t(x), C_g(x), N_t$}
	\KwOutput{$p^*_{s}(x), p^*_{t}(x), p, \lambda^*(x)$}
	Initialize $p^1_{s}(x), p^1_{t}(x), p^1, \lambda^1(x)$ \par
	\For{i = 1: $N_t$}
	{
		Optimizing $p$ by fixing other variables\
		$p^{i+1} = :\arg\max_{|p|\leq C_g(x)} -\frac{c}{2}|div p - p_s + p_t|^2 + \int_{\Omega}\lambda divpdx$ \par
		$p^{i+1} =: $Proj $[p^i + \alpha \nabla(divp - p_s + p_t)]_{|p^{i+1}(x)| \leq C_g(x)} $ \par
		Optimizing $p_s$ by fixing other variables\
		$p^{i+1}_{s} = :\arg\max_{p_s\leq C_s(x)} -\frac{c}{2}|divp p - p_s + p_t|^2 + \int_{\Omega}(1 -\lambda)p_sdx$ 
		
		$p^{i+1}_{s} = $Proj $[p^i_s + \alpha (divp + p_t- -p_s - (\lambda - 1) /c)]_{p^{i+1}(s) \leq C_s(x)} $ \par
		Optimizing $p_t$ by fixing other variables\
		$p^{i+1}_{t} = :\arg\max_{p_t\leq C_t(x)} -\frac{c}{2}|divp p - p_s + p_t|^2 + \int_{\Omega}\lambda p_tdx$ \
		
		$p^{i+1}_{t} = $Proj $[p^i_t + \alpha (- divp - p_t + p_s + \lambda /c)]_{p^{i+1}(s) \leq C_s(x)}$ \\
		Optimizing $\lambda$ by fixing other variables 
		$\lambda^{i+1} = :\arg\min_{\lambda} \int_{\Omega}\lambda(div p - p_s + p_t)dx$ \par
		$\lambda^{i+1} = : \lambda^{i} - \alpha (divp - p_s + p_t)$ \

	}
	$p^*_{s} : = p^{N_t + 1}_{s} $ ,
	$p^*_{t} : = p^{N_t + 1}_{t} $ ,
	$p^* : = p^{N_t + 1} $ , $\lambda^* : = \lambda^{N_t + 1} $
	\caption{ADMM Algorithm for Segmentation Inference}
\end{algorithm}


\subsection{Optimal Conditions for Max-flow/min-cut model}
The optimal condition for max-flow/min-cut model could be given by
\begin{align}
\label{eq7}
\lambda^*(x) = 0 &\implies p_s^*(x)  = C_s(x) \\
\label{eq8}
 \lambda^*(x) =1 &\implies p_t^*(x) = C_t(x) \\
 \label{eq9}
 |\nabla \lambda^*(x)|\neq 0 &\implies |p| = C_g(x)
\end{align} 
\begin{wrapfigure}{r}{0.55\textwidth}
	\centering
	\includegraphics[width=1\linewidth]{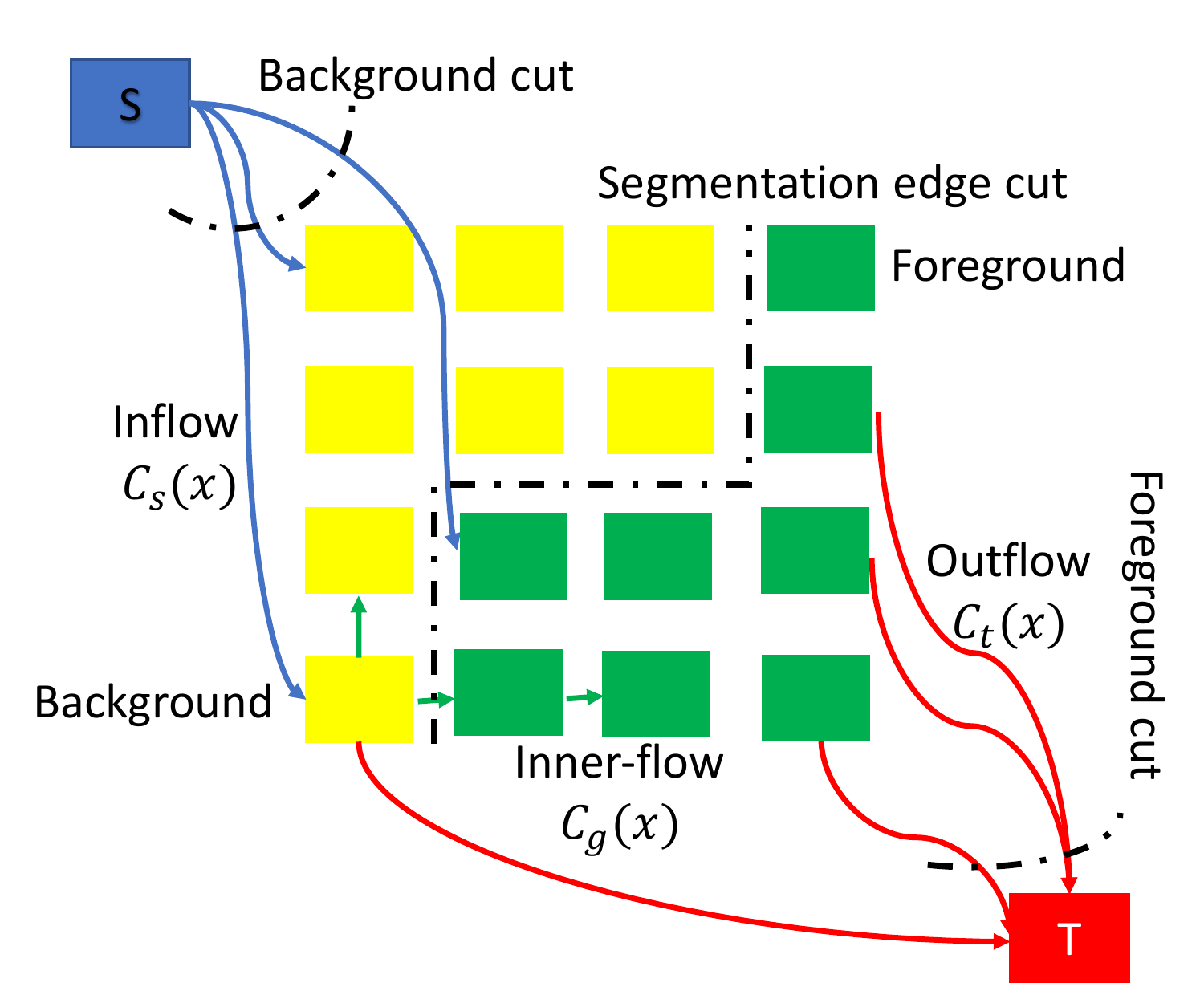}
	\caption{The max-flow/min-cut model for segmentation.}
	\vspace{-0.4in}
	\label{fig:flow}
\end{wrapfigure}
As it is shown in Fig. \ref{fig:flow}, an intuitive explanation for the optimal condition is that the incoming flow saturates at the optimal segmentation masks of background, the outgoing flow saturates at the optimal segmentation masks of foreground and the spatial flow saturates at the segmentation boundaries.
\subsection{Proposed Flow Regularized Training Losses}
\begin{figure}[htb]
	\centering
	\includegraphics[width=\linewidth]{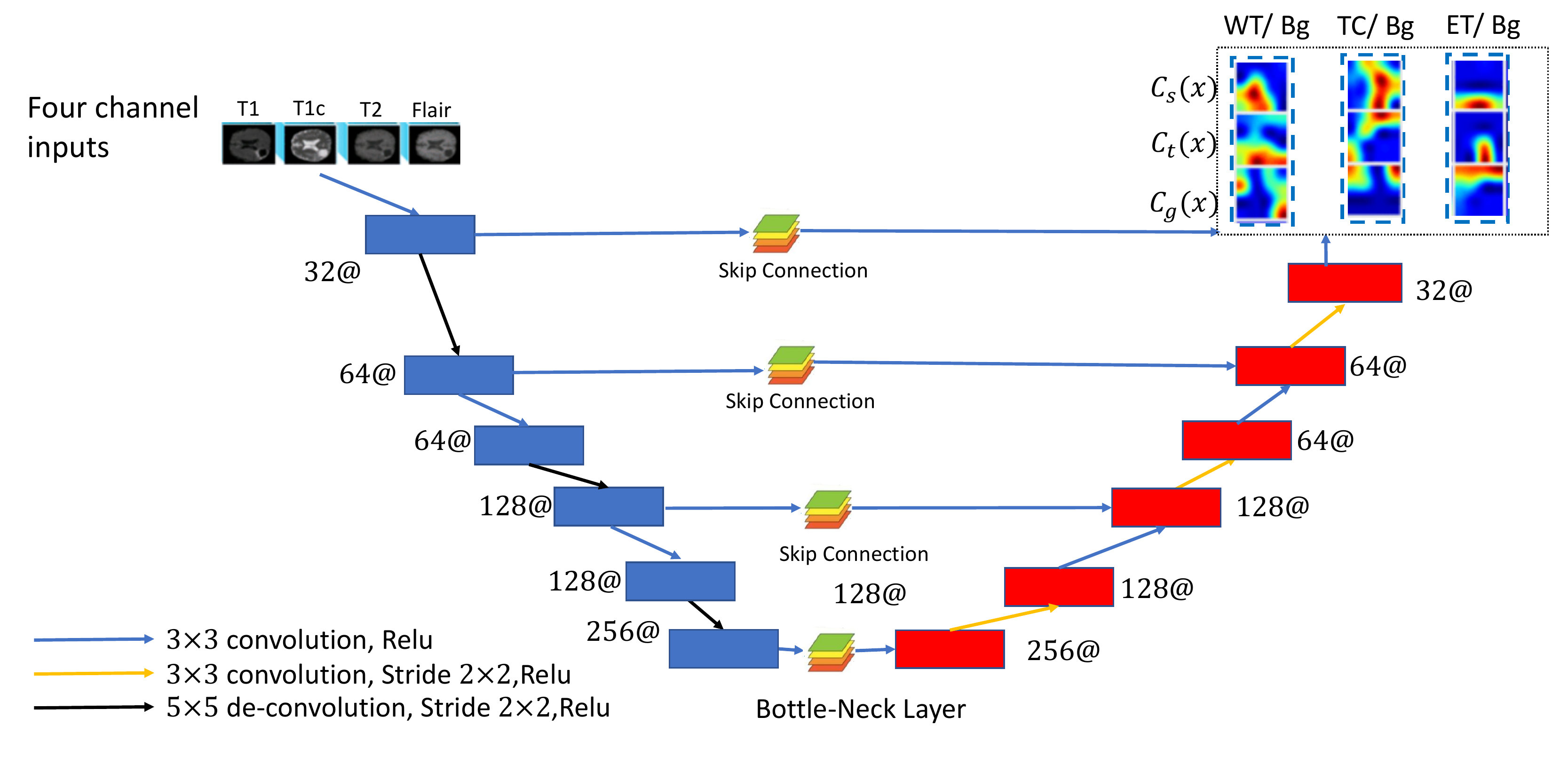}
	\caption{Our U-Net takes all four modalities MRI input and produces 9 feature maps output into 3 groups of $C_s(x)$ , $C_t(x)$ and $C_g(x)$.} \label{fig:arch}
\end{figure}

In our proposed methods, $C_s(x)$, $C_t(x)$ and $C_g(x)$ are taken from neural network output features rather than handcrafted features in traditional methods. Specifically, our $C_s(x)$, $C_t(x)$ and $C_g(x)$ come from the parameterized outputs of a deep U-Net as shown in Fig. \ref{fig:arch}.

Our U-Net takes the standardized input of four-modalities (T1, T2, Flair and T1c) MRI images. Then it passes through three down-sampling blocks. Each down-sampling block consists of a cascade of  a $3\times 3$ convolutional layer of stride 2 and another $3\times 3$ convolutional layer of stride 1 to reduce the size of feature map by half. The number of feature maps downward are 32, 64 ,64, 128, 128 and 256.  Following the bottle-neck layers, our U-Net takes three up-sampling blocks. Each up-sampling block consists of a cascade of  a $5\times 5$ deconvolutional layer of stride 2 and another $3\times 3$ convolutional layer of stride 1 to increase the size of feature map one time. Besides the feed-forward connections, our U-Net also consists skip connections between layers of the same horizon. Following the last layer of the deconvolutional block, the skip-connected feature maps are passing through the final $3\times 3$ convolutional layer of stride 1 to the final layer of 9 feature maps. These 9 feature maps are divided into 3 groups. Each group takes 3 feature maps as input of $C_s(x)$, $C_t(x)$ and $C_g(x)$. The three different groups are used to segment the whole tumor area, tumor core area and enhanced tumor core area from background hierarchically. \\
Recall that the saturated min-cut condition at optimal $\lambda^*(x)$ in (\ref{eq7}), (\ref{eq8}), ({\ref{eq9}}) and the flow conservation constraints (\ref{eq10}), we have the followings
\begin{equation}
\label{flow constraints}
 l_{\text{flow}} = |\hat{p}^*_s - p^*_s|  +  |\hat{p}^*_t - p^*_t| + |\hat{p}^* - p^*| =0 
\end{equation}
where $\hat{p}^*_s$, $\hat{p}^*_t$ and $\hat{p}^* $ are defined as 
\begin{equation}
\hat{p}^*_s   =
\begin{cases}
C_s(x)     & \quad \text{if} \hspace{0.5em} \lambda^*(x)  =0 \\
p^*_s  & \quad \text{if} \hspace{0.5em} \lambda^*(x) = 1
\end{cases}
\end{equation}
\begin{equation}
\hat{p}^*_t   =
\begin{cases}
C_t(x)     & \quad \text{if} \hspace{0.5em}\lambda^*(x)  =1 \\
p^*_t  & \quad \text{if} \hspace{0.5em}\lambda^*(x) = 0
\end{cases}
\end{equation}
\begin{equation}
\hat{p}^*   =
\begin{cases}
p^*     & \quad \text{if} \hspace{0.5em}|\nabla\lambda^*(x)|  =0 \\
p^* C_g(x) / |p^*| & \quad \text{if}\hspace{0.5em} |\nabla\lambda^*(x)|  \neq0 
\end{cases}
\end{equation}\\

The above equation holds as a result of joint optimal conditions of prime and duality. At the optimal points of primal variables $\lambda^*$ saturated flows at cutting edges equal to its maximum capacity constraints. And at the optimal points of
dual variables $p_s^*$, $p_t^*$ and $p^*$, the flow conservation function holds. By training on $l_{flow}$, we close both the primal and duality gaps at the point of ground truth segmentation $\lambda^*$. In our training function, we use Huber loss on $l_{flow}$.

\begin{figure}[htb]
	\centering
	\includegraphics[width=0.85\linewidth]{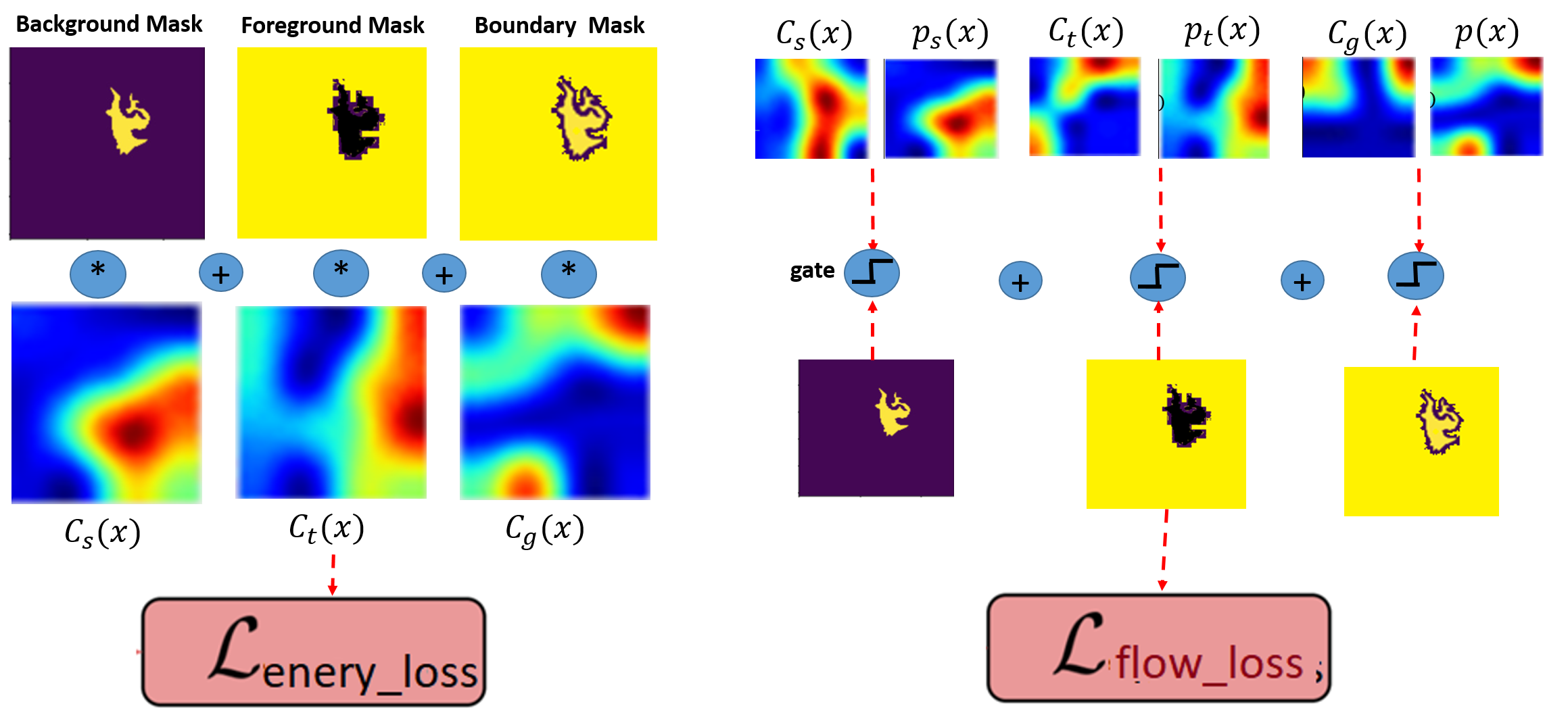}
	\caption{Our loss term includes two terms of energy loss and flow loss. The energy loss is in the same form as energy minimization function. The flow loss is in the form of primal-duality gaps between flows $p_s$, $p_t$ and $p$ and segmentation feature maps $C_g(x)$, $C_t(x)$ and $C_s(x)$. We enforce the constraints of optimal condition in the form of background mask on inflow $C_s(x)$ and $p_s$, foreground mask in outflow $C_t(x)$ and $p_t$ and boundary masks in edgeflow $C_g(x)$ and $p$. } \label{fig:loss}
\end{figure}
Our whole training loss function consists of an energy minimization term and a flow-regularized term as shown in Fig. \ref{fig:loss}.
\begin{equation}
\label{acmloss}
L_{\text{train}} = \underbrace{L_{\delta}(l_{\text{flow}})}_{\text{flow loss}} - \underbrace{\int_{\Omega}  \lambda^*(x)C_{s}(x)dx +\int_{\Omega}\lambda^*(x)C_{t}(x)dx + \int_{\Omega} C_g(x) |\nabla\lambda^*(x)|dx}_{\text{energy loss}}
\end{equation}
Our U-net is trained end-to-end by Algorithm \ref{al2}.\\
\begin{algorithm}[H]
	\label{al2}
	
	\KwInput{$\lambda^*(x)$, Image $\mathbf{I}$  , $\alpha$, $N_t$}
	\KwOutput{network parameters $\phi$}             
	
	Initialize $\phi^1$  \par

	\For{i = 1: $N_t$}
	{
		Feeding forward $\mathbf{I}$ \par
		$C_s(x) ,C_t(x) ,C_g(x) := f_{\phi^i}(\mathbf{I}) $ \par
		Running segmentation inference algorithm\par
		$p^*_s(x), p^*_t(x), p^*_g(x) : =  \text{Infer} (C_s(x) ,C_t(x) ,C_g(x)) $ \par
		Getting flow regularization loss from segmentation label \par
		 $L_{\delta}(\hat{p}^*_s + -  \hat{p}^*_t - div\hat{p}^*)$ \par
		Getting energy minimization loss from segmentation label \par 
		$  \int_{\Omega} - \lambda^*(x)C_{s}(x)dx
		+\int_{\Omega} \lambda(x)^*C_{t}(x)dx + \int_{\Omega} C_g(x) |\nabla\lambda^*(x)|dx$ \par 
		Updating $\phi$ from loss gradient \par
		$\phi^{i+1} :=\phi^{i} - \alpha \nabla(L_{train})  $

	}
	$\phi : = \phi^{N_t + 1} $ \
	\caption{U-Net Training Algorithm}
\end{algorithm}

%% file: exp.tex
\section{Experiments} 
\textbf{Experiment Settings} We evaluate our proposed methods in BRATS2018~\cite{menze2014multimodal} dataset and compare it with other state-of-the-art methods. We randomly split the dataset of 262 glioblastoma scans of 199 lower grade glioma scans into training ($80\%$) and testing ($20\%$).  We evaluate our methods following the BRATS challenge suggested evaluation metrics of Dice, Sensitivity (Sens) and Specificity (Spec) and Hausdorff. And we report our segmentation scores in three categories of whole tumor area (WT), tumor core (TC) and enhance tumor core (EC).
\begin{table}
	\centering
	\caption{Segmentation results in the measurements of Dice score, Sensitivity and Specificity.}
	\begin{tabular}{| c | c  c  c | c c c| c c c|}
		\hline
		&   & \textbf{Dice Score} &  &   & \textbf{Sensitivity} & &   & \textbf{Specificity} & \\ \hline
		&  WT \hspace*{0.5em} &  TC\hspace*{0.5em} &  EC\hspace*{0.5em} &  WT \hspace*{0.5em} &  TC\hspace*{0.5em} &  EC\hspace*{0.5em}&  WT \hspace*{0.5em} &  TC\hspace*{0.5em} &  EC\hspace*{0.5em}
		\\ \hline
		Deep Medic~\cite{kamnitsas2015multi}  &0.896  & 0.754  & 0.718 &0.903  & 0.73  & 0.73& N/A  & N/A  & N/A
		\\ \hline
		DMRes \cite{kamnitsas2017efficient}  & \textbf{0.896}   &  0.763  &  0.724 &\textbf{0.922}  & 0.754  & 0.763 & N/A  & N/A  & N/A
		\\ \hline
		DRLS \cite{le2018deep} & 0.88   &  0.82  &  0.73 &0.91  & 0.76  & 0.78 &0.90  & 0.81  & 0.71

		\\ \hline
		Proposed  & 0.89  & \textbf{0.85}  &  \textbf{0.78} & 0.92  & \textbf{0.79}  & \textbf{0.78} &\textbf{0.93}  & \textbf{0.83}  & \textbf{0.75}
		\\ \hline	
	\end{tabular}
	\label{table:perf}
\end{table}

\noindent\textbf{Implementation Details}
In training phase, we use a weight decay of $1e-6$ convolutional kernels with a drop-out probability of $0.3$. We use momentum optimizer of learning rate 0.002. The optimal dual $p_s$, $p_t$ and $p$ used in our training are instantly run from 15 steps of iterations with a descent rate of $\alpha =0.16$ and $c=0.3$. In our quantitative evaluation, we empirically select to use the 15th iteration result $\lambda^{15}(x)$ and thresholds it with $l=0.5$.


	

\begin{table}
	\centering
	\caption{We report result of our proposed methods with ACM loss in the lower section of our table and comparing it with the result of the baseline without ACM loss the uppper section of our table.}
	\begin{tabular}{| c | c  c  c | c c c| }
	
		\hline
		&  & \textbf{Dice Score} &  &   & \textbf{Hausdorff} &  \\ \hline
		w/o ACM &  WT \hspace*{0.5em} &  TC\hspace*{0.5em} &  EC\hspace*{0.5em} &  WT \hspace*{0.5em} &  TC\hspace*{0.5em} &  EC\hspace*{0.5em}
		\\ \hline
		Mean &0.87  & 0.82  & 0.74 &5.03 & 9.23 & 5.58
		\\ \hline
		Std   & 0.13   &  0.18  &  0.26 &6.54  & 11.78  & 6.27 
		\\ \hline
		Medium & 0.89   &  0.87  &  0.83 &4.78  & 8.59  & 4.92 

		\\ \hline
		25 Quantile  & 0.86  & 0.77  &  0.73 & 4.23  & 8.12  & 4.21
		\\ \hline	
		75 Quantile  & 0.95  & 0.88  &  0.85 & 4.91  & 8.87  & 5.24
		\\ \hline
		w/ ACM &  WT \hspace*{0.5em} &  TC\hspace*{0.5em} &  EC\hspace*{0.5em} &  WT \hspace*{0.5em} &  TC\hspace*{0.5em} &  EC\hspace*{0.5em}
		\\ \hline
		Mean & \textbf{0.89}   & \textbf{0.85}  & \textbf{0.78} & \textbf{4.52}  & \textbf{6.14}  & \textbf{3.78}
		\\ \hline
		Std   & 0.12   &  0.21  &  0.28 &5.92  & 7.13  & 4.18 
		\\ \hline
		Medium & 0.91   &  0.88  &  0.77 &2.73  & 3.82  & 3.13 

		\\ \hline
		25 Quantile  & 0.87  & 0.83  &  0.75 & 1.83  & 2.93  & 2.84
		\\ \hline	
		75 Quantile  & 0.93  & 0.90  &  0.80 & 3.18  & 5.12  & 3.52
		\\ \hline
	\end{tabular}
	\label{table:acm}
\end{table}
	





\subsection{Quantitative Results}
A quantitative evaluation of results obtained from our implementation of proposed methods is shown in Table \ref{table:perf}.  The experiment results show that our performances are comparable with state-of-the-art results in the categories of all metrics of sensitivity, specificity and dice score. We perform data ablation experiments by substituting our ACM in Eq.\ref{acmloss} with standard cross-entropy loss. The comparison of our proposed methods with the one without ACM loss is shown in Table \ref{table:acm}. 
The trainable active contour model would increase performance on the same U-Net structure. 

\vspace{-1em}
\begin{figure}[htb]
	\centering
	\includegraphics[width=0.9\linewidth]{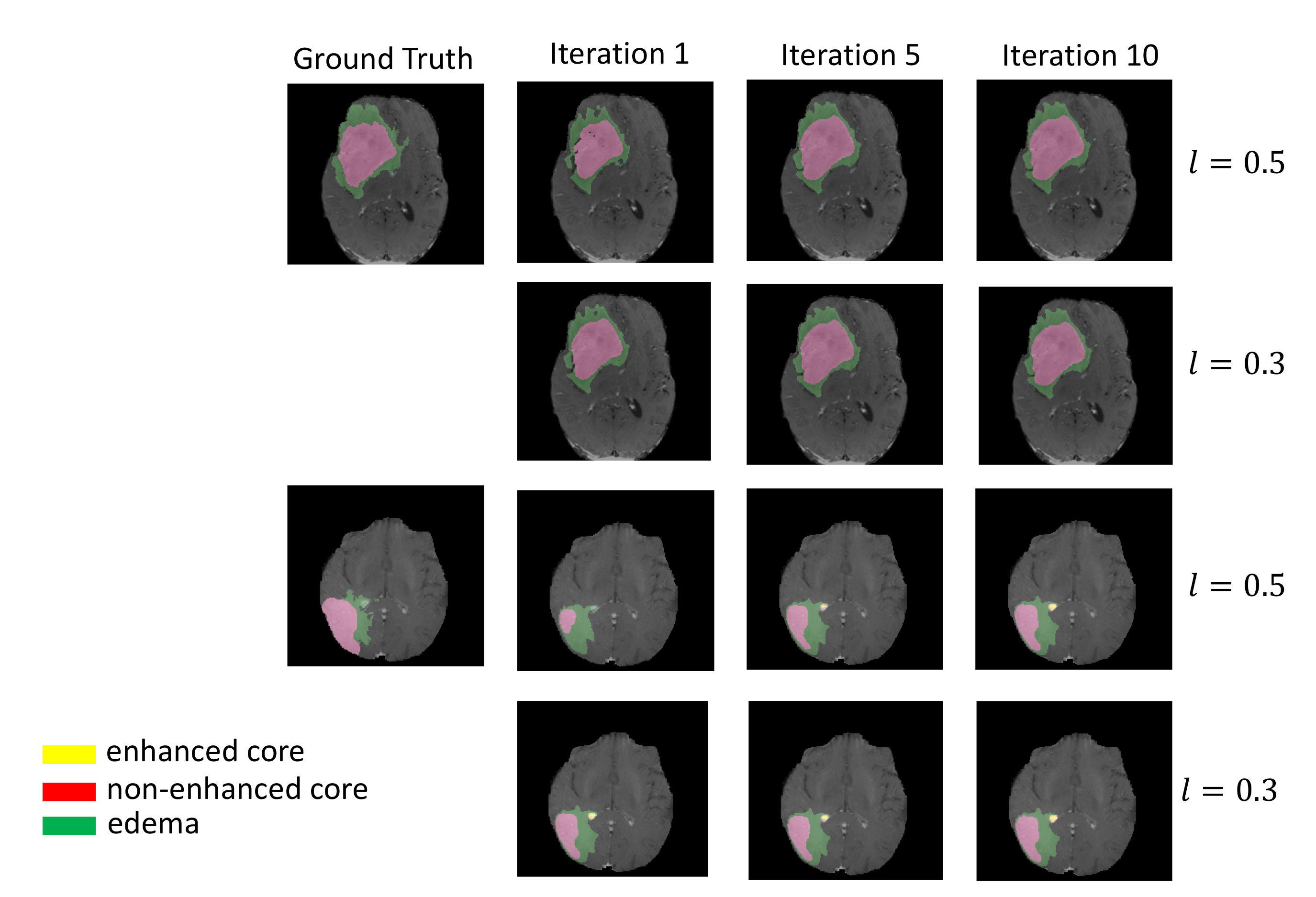}
	\caption{Segmentation results with various number of iterations and level set thresholds.} \label{fig:quali}
\end{figure}
\vspace{-1em}
\subsection{Qualitative Results}
Fig. \ref{fig:quali} shows the active contours evolving with different number of iterations and different level set thresholds. The figure shows two examples with one iteration, five iterations, and ten iterations, and with level set thresholds of 0.5 and 0.3, respectively. Increasing the number of iterations tends to have smoothing effects of the boundaries and filtering outlying dots and holes. Changing the level set threshold values would cause a trade-off between specificity and sensitivity. The combination of deep learning models and active contour models provides the flexibility to adjust the results.